# Fractional-order spike-timing-dependent gradient descent for multi-layer spiking neural networks

Yi Yang, Richard M. Voyles, Haiyan H. Zhang, Robert A. Nawrocki*

*School of Engineering Technology, Purdue University, West Lafayette, United States of America*

## Abstract

Accumulated detailed knowledge about the neuronal activities in human brains has brought more attention to bio-inspired spiking neural networks (SNNs). In contrast to non-spiking deep neural networks (DNNs), SNNs can encode and transmit spatiotemporal information more efficiently by exploiting biologically realistic and low-power event-driven neuromorphic architectures. However, the supervised learning of SNNs still remains a challenge because the spike-timing-dependent plasticity (STDP) of connected spiking neurons is difficult to implement and interpret in existing backpropagation learning schemes. This paper proposes a fractional-order spike-timing-dependent gradient descent (FO-STDGD) learning model by considering a derived nonlinear activation function that describes the relationship between the quasi-instantaneous firing rate and the temporal membrane potentials of nonleaky integrate-and-fire neurons. The training strategy can be generalized to any fractional orders between 0 and 2 since the FO-STDGD incorporates the fractional gradient descent method into the calculation of spike-timing-dependent loss gradients. The proposed FO-STDGD model is tested on the MNIST and DVS128 Gesture datasets and its accuracy under different network structure and fractional orders is analyzed. It can be found that the classification accuracy increases as the fractional order increases, and specifically, the case of fractional order 1.9 improves by 155% relative to the case of fractional order 1 (traditional gradient descent). In addition, our scheme demonstrates the state-of-the-art computational efficacy for the same SNN structure and training epochs.

## 1. Introduction

Artificial neural networks (ANNs) have witnessed their wide applications in many aspects of modern science and technology, including tackling the image and signal processing tasks, recommender system development, object detection , and solving the robot motion planning problems [1,2]. As the third-generation ANNs, spiking neural networks (SNNs) have attracted sustained and extensive attention from the artificial intelligence community due to their biological plausibility, computational efficiency, and low power consumption in both the software simulation and hardware emulation [3–13]. In SNNs, distinct from their non-spiking ANNs, the encoding of information is accomplished through the precise timing of a series of asynchronous action potentials—commonly referred to as spikes—that traverse the network of interconnected neurons. This mode of operation eschews reliance on the shape and amplitude of electrical analogue potentials, favoring a mechanism that ensures efficient and low-power signal processing [13,14].

Within the neuromorphic computing community, there prevails a consensus that synaptic plasticity in SNNs—characterized by alterations in synaptic weights/efficacy—is predominantly orchestrated by an unsupervised learning protocol predicated on the inherent self-correlation of spike timing, a phenomenon referred to as spike-timing dependent plasticity (STDP) [15]. Concurrently, neuroscientific study has illuminated that supervised learning, a critical framework within non-spiking ANNs, also possesses a biologically plausible counterpart in the form of STDP within SNNs, thus intimating at the expansive utility and integration of this learning paradigm across computational models [13,15–17]. However, the formulation of bio-plausible supervised learning models for SNNs remains a challenging problem due to the intricate spatiotemporal behavior of spiking neurons and the inherently non-differentiable nature of discrete spiking events. Consequently, many high-performance gradient-based supervised learning algorithms, such as the steepest descent method, stochastic gradient descent, and fractional gradient

———
* Corresponding author.
*E-mail addresses:* yang1087@purdue.edu (Y. Yang), rvoyles@purdue.edu (R.M. Voyles), hhzhang@purdue.edu (H.H. Zhang), robertnawrocki@purdue.edu (R.A. Nawrocki)

decent, which have been effectively applied in non-spiking ANNs, are not readily transferable to SNNs [18–21].

In recent decades, a substantial body of supervised learning algorithms for SNNs has emerged, which can largely be categorized into two distinct groups. The first group encompasses algorithms that utilize a pretrained ANN to approximate the overall neuronal dynamics of the SNN through meticulously calibrated hyperparameters [22–26]. However, these ANN-inspired SNNs often exhibit diminished representational and learning capabilities relative to their ANN equivalents, primarily due to suboptimal exploitation of spike timing information during the translation process from ANNs to SNNs.

The second group includes algorithms that rely on the continuous approximation (CA) of the membrane potential in close proximity to the neuron's firing instances [27–37]. This CA may be characterized by either a linear or nonlinear function, or alternatively, tackled through continuous surrogate gradient methods [33,38]. Notably, SpikeProp [27], an early supervised learning algorithm tailored for SNNs, postulates that the membrane potential can be approximated as a linear function around the firing times. Subsequent iterations of SpikeProp have introduced varied network architectures and coding strategies to enhance learning stability and computational efficiency [28,39,40]. A particularly significant development has been the introduction of the time-to-first-spike (TTFS) coding scheme, which has demonstrated the potential to transform the non-differentiable spikes of nonleaky Integrate-and-Fire (IF) neurons into an input-output relationship that is differentiable almost everywhere [41]. Furthermore, building on the concept of locally exact derivatives concerning the pre- and postsynaptic spike timings, Comsa et al. [34] have formulated a continuously differentiable spike timing function expressed through the Lambert W function [42].

While SNNs trained with CA algorithm exhibit enhanced performance over their ANN-derived counterparts, several challenges persist, including issues such as inactive (dead) neurons, gradient exploding, and a departure from biological plausibility, which necessitates supervised learning models that incorporates a degree of spike-timing dependency [41,43,44]. Despite a few efforts to align STDP with supervised learning in SNNs [45,46], no existing study has successfully integrated biologically plausible spike-timing dependency into gradient-based supervised learning models. It is this gap that motivates our current endeavor to develop a SNN training model that accommodates such principles.

In this paper, we present a novel fractional-order spike-timing-dependent gradient descent (FO-STDGD) learning model that capitalizes on the principles of fractional gradient descent. Our contributions are twofold: (1) We provide a rigorous proof of the convergence of the FO-STDGD model to the real minimum of the energy function. This analytical proof lays a solid theoretical foundation, affirming the viability of our algorithm and its potential to facilitate superior learning performances in SNN environments; (2) Through meticulous comparative analysis with existing SNN learning models, we offer evidence that the FO-STDGD model maintains a competitive edge in terms of classification accuracy and computational efficiency, suggesting that our approach is a noteworthy contribution to the field of neuromorphic computing.

## 2. Method

Prior to developing our spike-based supervised learning rules, we adopt the nonleaky Integrate-and-Fire (IF) model to describe single neuron dynamics and demonstrate how the temporal activity (quasi-instantaneous firing rate) of a single neuron is related to its mean membrane potential at firing times. The relationship (or the activation function), that is established and justified in Section 2.1, initializes our steps for transforming the backpropagation-based learning rules that update localized states in conventional non-spiking neural networks to spike timing dependent plasticity of synapses in spiking neural networks. In addition, the derivation of the FO-STDGD model relies on the application of fractional gradient descent method in Section 2.2, because the fractional gradient descent method utilizes the nonlocal and hereditary properties of fractional-order derivatives to reinforce the spike-timing dependency of the novel synaptic updating rules. The complete FO-STDGD model is then derived in Section 2.3 by fully employing the activation function and the fractional gradient descent method.

### 2.1. Nonleaky IF neuron and its activation function

In the context of a nonleaky IF neuron, the membrane potential undergoes incremental growth as it accumulates the effects of incoming presynaptic spikes within the membrane capacitor. The discretized nonleaky IF neuron model presumes that the neuron can only emit action potentials (spikes) at discrete time instances, specifically at $t = k\Delta t, k \in \mathbb{N}$ where $k$ is a natural number and $\Delta t = 1$ ms represents the simulation time step size (For simplicity, $\Delta t$ is assumed to be a unity factor and omitted in the subsequent discrete-time models and relevant derivations). The forward propagation of two fully connected layers of SNNs is depicted in Fig. 1, with the membrane potential of the $i^{\text{th}}$ neuron in the $l^{\text{th}}$ layer at time $t$ being expressed as follows,

$$u_i^{[l]}(t) = u_i^{[l]}(t-1) + w_i^{[l]}s^{[l-1]}(t) \tag{1}$$

where $u_i^{[l]}(t)$ and $u_i^{[l]}(t-1)$ represent the membrane potentials at two consecutive discrete time points, respectively. $w_i^{[l]}(t) = [w_{i1}^{[l]}(t), w_{i2}^{[l]}(t), \ldots, w_{in^{[l-1]}}^{[l]}(t)] \in \mathbb{R}^{n^{[l-1]}}$ denotes the synaptic weights, $s^{[l-1]}(t) = [s_1^{[l-1]}(t), s_2^{[l-1]}(t), \ldots, s_{n^{[l-1]}}^{[l-1]}(t)]^T \in \mathbb{R}^{n^{[l-1]}}$ signifies the output spike vector for $n^{[l-1]}$ neurons in the $(l-1)^{\text{th}}$ layer at time $t$ with $s_i^{[l-1]}(t) \in \{0,1\}$. When the membrane potential $u_i^{[l]}(t)$ reaches the firing threshold voltage $\theta$, the postsynaptic spike is promptly generated, and simultaneously, $u_i^{[l]}(t)$ is reset to the resting potential (assume $u_{rest} = 0$). In this context, if we represent the firing time of the $i^{\text{th}}$ neuron in the $l^{\text{th}}$ layer as $t_i^{[l],f} \in \{t \in \mathbb{N} | s_i^{[l]}(t) = 1\}$, the firing condition for the $i^{\text{th}}$ neuron can be articulated as

$$\left(u_i^{[l]}(t_i^{[l],f,-}) \geq \theta\right) \bigwedge \frac{\mathrm{d}u_i^{[l]}(t)}{\mathrm{d}t}\bigg|_{t=t_i^{[l],f,-}} > 0 \tag{2}$$

where $\bigwedge$ is the logic "and" operator, and $t_i^{[l],f,-}$ represents the pre-instantiate of the firing time $t_i^{[l],f}$. Note that if (2) is satisfied, the neuron fires a spike $s_i^{[l]}\left(t_i^{[l],f}\right) = 1$ and membrane potential is reset $u_i^{[l]}(t_i^{[l],f,+}) = u_{rest}$, where $t_i^{[l],f,+}$ denotes the post-instantiate of the firing time $t_i^{[l],f}$.

Fig. 1 illustrates the distinctions between our multi-layer-perceptron (MLP) based SNN and traditional non-spiking ANN. Unlike traditional non-spiking neurons with lumped parameters and localized states, the spiking neurons in our study encode spatiotemporal action potentials individually in the discrete-time domain but analyse its error propagation statistically in the continuous-time domain. As nonleaky IF neuron model (1) is described in the discrete time space, the generated spike trains can be expressed using the Kronecker delta function, where $\delta(t, t_i^{[l],f}) = 1$ if $t = t_i^{[l],f}$ and otherwise, $\delta(t, t_i^{[l],f}) = 0$. However, to transition the nonleaky IF model from the discrete-time domain to the continuous-time domain, we intentionally propose the temporal average membrane potential $\hat{u}_i^{[l]}(t)$ and the quasi-instantaneous firing rate $\hat{s}_i^{[l]}(t)$ by averaging the membrane potentials and action potentials (spikes) within a pre-selected time window $\tau$,

$$\hat{u}_i^{[l]}(t) = \frac{1}{\tau} \sum_{t_i^{[l],f} \in (t-\tau, t]} u_i^{[l]}(t_i^{[l],f}) \tag{3}$$

$$\hat{s}_i^{[l]}(t) = \frac{1}{\tau} \sum_{t_i^{[l],f} \in (t-\tau, t]} s_i^{[l]}\left(t_i^{[l],f}\right) = \frac{1}{\tau} \sum_{s \in (t-\tau, t]} s_i^{[l]}(s) \tag{4}$$

where the two summations on the right side of Equation (4) are equal because $s_i^{[l]}(t) = 1$ for $t = t_i^{[l],f}$ and $s_i^{[l]}(t) = 0$ otherwise. The time window $\tau$ is a characteristic parameter that statistically regulates the temporal activity of spiking neurons in the continuous-time domain, and $\tau$ is chosen from $\{\tau \in \mathbb{N} | 1 < \tau \leq T\}$, where $T$ is the length of the simulation window for training the SNNs. Moreover, synaptic homeostasis plays a vital role in preserving stable and uniform computational dynamics across different neurons within the same network layer. We assert that $\|w_i^{[l]}(t)\|_1 < \theta$ for all $l$ and $t \in (0, T)$, which implies that synaptic homeostasis prevents abrupt changes in membrane potential within a single simulation time step and ensures the refractoriness of each spiking neuron. Consequently, in the forward propagation pathway of multi-layer SNNs, we can show the existence of a nonlinear (piecewise linear) activation function that establishes the relationship between $\hat{s}_i^{[l]}(t)$ and $\hat{u}_i^{[l]}(t)$ in this study.

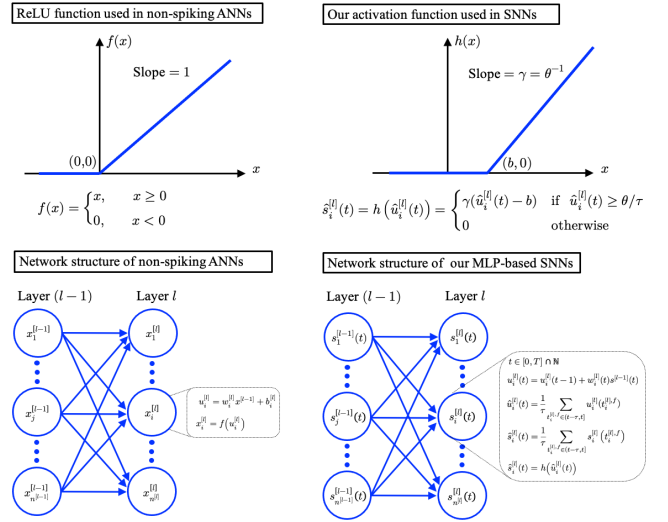

Fig. 1. Comparative visualization of the activation functions and structural configurations distinguishing traditional non-spiking artificial neural networks from our proposed spiking neural network model.

**Theorem 1.** The temporal average membrane potential and the quasi-instantaneous firing rate of a nonleaky IF neuron satisfies the activation function $\hat{s}_i^{[l]}(t) = h(\hat{u}_i^{[l]}(t))$, i.e.,

$$\hat{s}_i^{[l]}(t) = \begin{cases} \gamma(\hat{u}_i^{[l]}(t) - b) & \text{if} \quad \hat{u}_i^{[l]}(t) \geq \theta/\tau \\ 0 & \text{otherwise} \end{cases} \tag{5}$$

where $\gamma = \theta^{-1}$, $0 \leq b < \theta$, and $\theta$ denotes the threshold voltage of the nonleaky IF neuron.

**Proof.** In a simple case, assume that there is only one spike generated by the $i^{\text{th}}$ postsynaptic neuron in the $l^{\text{th}}$ layer within $(t-\tau, t]$, and suppose that the firing time $t_i^{[l],f}$ is exactly at $t$. Thus, $\hat{s}_i^{[l]}(t) = 1/\tau$. Plus, (2) and (3) implies that $\hat{u}_i^{[l]}(t) \geq \theta/\tau$. We then define $b = \hat{u}_i^{[l]}(t) - \hat{s}_i^{[l]}(t)/\gamma$ for $\gamma = \theta^{-1}$ and immediately have $b \geq 0$. Thereafter, in order to show that $b < \theta$, we can draft a proof by contradiction. Suppose that $b \geq \theta$, then one has $\hat{u}_i^{[l]}(t) = \hat{s}_i^{[l]}(t)/\gamma + b \geq \theta/\tau + \theta = (\tau+1)\theta/\tau$. And from (3), one has $\sum_{t_i^{[l],f} \in (t-\tau, t]} u_i^{[l]}(t_i^{[l],f}) \geq (\tau+1)\theta$. Thus, the membrane potential at the firing time should satisfy $u_i^{[l]}(t) > (\tau+1)\theta > 2\theta$. The synaptic homeostasis and (1) together implies that $u_i^{[l]}(t-1) \geq u_i^{[l]}(t) - \|w_i^{[l]}(t)\|_1 > \theta$, which means that the neuron fires at time $t-1$, which is a contradiction to our assumption. Furthermore, according to the pigeonhole principle, we can have another interesting observation: $\sum_{s \in (t-\tau, t-1]} u_i^{[l]}(s) < (\tau-1)\theta$. Note that $\tau \in \mathbb{N}$ and $\tau > 1$. If $\sum_{s \in (t-\tau, t-1]} u_i^{[l]}(s) \geq (\tau-1)\theta$, the membrane potential will increase no less than $\theta$ within at least one simulation time step. And since $u_i^{[l]}(s) \geq u_{rest} = 0$ forever, we must observe at least one spike generated by the neuron within $(t - \tau, t - 1]$, which is a contradiction.

In a more general case, assume that there are multiple spikes generated within $(t - \tau, t]$. Without loss of the generality, we suppose that one spike is generated exactly at time $t$. Similarly, if we assume that $b \geq \theta$, we can have $\sum_{t_i^{[l],f} \in (t - \tau, t]} u_i^{[l]}(t_i^{[l],f}) \geq (\tau + 1)\theta$. Since at most $\tau$ spikes can be generated within $(t - \tau, t]$, according to the pigeonhole principle, we must have at least one firing time, say $t^f$, at which the membrane potential is greater than $2\theta$, which implies that $u_i^{[l]}(t^f - 1) = u_i^{[l]}(t^f) - \|w_i^{[l]}(t^f)\|_1 > \theta$ and the neuron fires at $t^f - 1$. This leads to a contradiction because $u_i^{[l]}(t^f - 1)$ should be immediately reset to 0 once a spike is generated at $t^f - 1$. □

**Remark 1.** We introduce a nonlinear activation function for spiking neurons, facilitating their training through both forward and backward propagation. This unique approach encodes spike patterns for individual neurons by averaging membrane potential and spike counts, reinforcing synaptic plasticity in alignment with Hebb's postulate: "neurons that fire together, wire together" [47]. Furthermore, we adopt $\hat{s}_i^{[l]}(t)$ (for $\tau > 1$) as opposed to $s_i^{[l]}(t)$. $\hat{s}_i^{[l]}(t)$ can assume continuous values within the range $[0,1]$, whereas $s_i^{[l]}(t)$ is binary, taking values of $\{0,1\}$. This 'continuousization' imparts to Equation (5) properties that are characteristics of rectified linear unit (ReLu) function. The piecewise linear nature of our novel activation function, akin to ReLu, contributes to its simplicity and computational efficiency, which are highly conducive to hardware implementations.

### 2.2. Fractional gradient descent method

The fractional gradient descent method serves as a generalization to the traditional gradient descent method [20,48–50]. The $\alpha$-th order Caputo's fractional derivative of $f(t)$ is defined as

$$_{t_0}^{C}D_t^\alpha f(t) = \frac{1}{\Gamma(n - \alpha)} \int_{t_0}^{t} \frac{f^{(n)}(s)}{(t - s)^{\alpha - n + 1}} ds \quad (6)$$

where $n - 1 < \alpha \leq n, n \in \mathbb{N}^+$, $\Gamma(n - \alpha)$ is the Gamma function, and $t_0$ is the initial time of the integration. If $f \in C^{n-1}([t_0, t])$ and $f^{(n)} \in L^1([t_0, t])$, (6) can be expressed as an infinite series [50],

$$_{t_0}^{C}D_t^\alpha f(t) = \sum_{i=n}^{\infty} \frac{f^{(i)}(t_0)}{\Gamma(i + 1 - \alpha)} (t - t_0)^{i - \alpha} \quad (7)$$

Consider that $f$ is a convex function defined in a compact set $G$, the contraction mapping theorem demonstrates the existence of a unique extreme point denoted as $x^*$ in $G$. This extreme point can be obtained iteratively, through the conventional (integer-order) gradient descent method, i.e., $x_{k+1} = x_k - \mu f^{(1)}(x_k)$, where $k$ is the iteration count and $\mu$ is the learning rate. Nevertheless, deviating from the traditional gradient descent approach involves the replacement of the standard first derivative with Caputo's fractional derivative, leading to the fractional gradient descent method [20,48], i.e., $x_{k+1} = x_k - \mu \cdot {}_{x_0}^{C}D_{x_k}^\alpha f(x_k)$. It is noteworthy, however, that this expression encounters challenges in converging toward the actual/real extremum of $f$. The divergence is attributed to the nonlocal nature of the operator ${}_{x_0}^{C}D_{x_k}^\alpha$, which exhibits a strong dependency on the initial value $f(x_0)$. Consequently, this dependency propels the sequence towards a spurious extremum, a concept defined within the context of fractional derivatives [48]. To address the intricacies associated with this ill-posed convergence trajectory, a refined fractional gradient descent method was introduced in [20] and rephrased as Equation (8) in this study. This revised formula effectively mitigates the impact of the nonlocal fractional operator by reducing the size of the action interval,

$$x_{k+1} = x_k - \mu \cdot {}_{x_{k-1}}^{C}D_{x_k}^\alpha f(x_k) \quad (8)$$

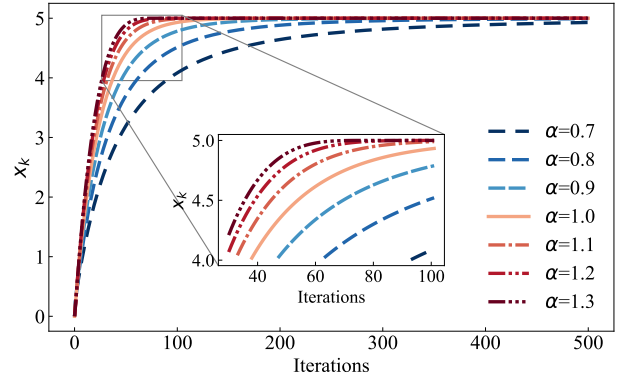

Fig. 2. Comparison of the convergence speed for fractional gradient descent method under different $\alpha$ values.

Upon substituting Equation (7) into Equation (8) and retaining solely the first term, an iterative formula can be obtained,

$$x_{k+1} = x_k - \mu \frac{f^{(1)}(x_{k-1})}{\Gamma(2 - \alpha)} |x_k - x_{k-1} + \varepsilon|^{1-\alpha} \quad (9)$$

where $0 < \alpha < 2$ is required to ensure that the approximation of fractional gradient operator can be interpreted as an extension of the conventional gradient operator ($\alpha = 1$). Additionally, $\varepsilon$ is introduced as a small value to circumvent potential singularities arising from $x_k$ being equal to $x_{k-1}$. The use of the absolute value within the stepwise difference serves to enhance computational stability. Furthermore, it is noteworthy that a rigorous proof of the convergence of Equation (9) towards the real extremum of a convex function has been provided in a previous work [51][2].

To assess the convergence rates of the fractional gradient descent method across different $\alpha$ values, we examine an objective function $f(x) = (x - 5)^2$ and employ Equation

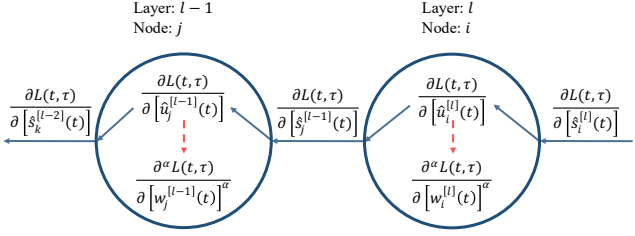

Fig. 3. The backpropagation path of two neurons in two connected layers.

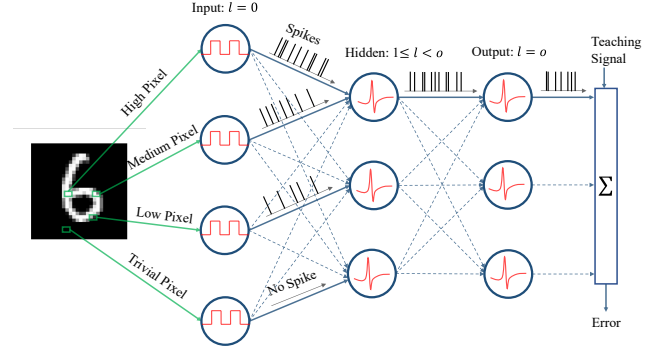

Fig. 4. The architecture of a multi-layer spiking neural network for image classification task, consisting of an input layer that converts image pixel intensity to Poisson spike trains, hidden layers and an output layer .

(9) to determine its minimum. The simulation results, using a learning rate of 0.02, are depicted in Fig. 2. Notably, our findings reveal an enhanced convergence speed as the fractional order $\alpha$ increases. Furthermore, when $\alpha$ surpasses 1, the fractional gradient descent method demonstrates superior performance compared to the conventional gradient descent method. These observations suggest that the forthcoming Fractional-Order Spike-Timing-Dependent Gradient Descent (FO-STDGD) method, as presented in the subsequent section, has the potential to outperform traditional gradient descent-based or Spike-Timing-Dependent Plasticity (STDP)-based learning rules.

## 2.3. FO-STDGD learning rule

The initial stage in the training process of our multi-layer SNNs entails the forward propagation of both the instantaneous firing rate and the average membrane potential for each neuron, following the principles outlined in Equations (10) and (5),

$$\hat{u}_i^{[l]}(t) = \hat{u}_i^{[l]}(t-1) + w_i^{[l]}(t)\hat{s}^{[l-1]}(t) \tag{10}$$

Given that the spatiotemporal patterns for individual neurons can be forward-propagated through a nonlinear activation function, the synaptic weights are updated by backward-propagating the errors from the output layer to the input layer. In Fig. 3, it is evident that the integer-order gradients, connecting the loss function $L(t,\tau)$ with the nonlocal states of each neuron, can be directly calculated and relayed through the application of the conventional chain rule. However, it is crucial to underscore that the fractional derivatives of the loss function concerning the synaptic weights are computed within each neuron and do not extend to the preceding layers. Correspondingly, the backward-propagation process can be described as

$$\frac{\partial L(t,\tau)}{\partial[\hat{u}_i^{[l]}(t)]} = \frac{\partial L(t,\tau)}{\partial[\hat{s}_i^{[l]}(t)]}\frac{\partial[\hat{s}_i^{[l]}(t)]}{\partial[\hat{u}_i^{[l]}(t)]} = \frac{\partial L(t,\tau)}{\partial[\hat{s}_i^{[l]}(t)]}h^{(1)}(\hat{u}_i^{[l]}(t)) \tag{11}$$

$$\frac{\partial L(t,\tau)}{\partial[\hat{s}_j^{[l-1]}(t)]} = \sum_{i=1}^{n^{[l]}}\frac{\partial L(t,\tau)}{\partial[\hat{s}_i^{[l]}(t)]}\frac{\partial[\hat{s}_i^{[l]}(t)]}{\partial[\hat{u}_i^{[l]}(t)]}\frac{\partial[\hat{u}_i^{[l]}(t)]}{\partial[\hat{s}_j^{[l-1]}(t)]}$$
$$= \sum_{i=1}^{n^{[l]}}\frac{\partial L(t,\tau)}{\partial[\hat{u}_i^{[l]}(t)]}w_{ij}^{[l]}(t) \tag{12}$$

where $h^{(1)}()$ denotes the first derivative of the activation function defined in (5). The second equal signs in Equations

(11) and (12) are derived from the relationships established in Equations (5) and (1), respectively. In addition, the synaptic weights contingent upon spike-timing are subject to

$$\frac{\partial^\alpha L(t,\tau)}{\partial[w_{ij}^{[l]}(t)]^\alpha} = \frac{\partial L(t,\tau)}{\partial[\hat{u}_i^{[l]}(t)]}\frac{\partial^\alpha[\hat{u}_i^{[l]}(t)]}{\partial[w_{ij}^{[l]}(t)]^\alpha}$$
$$= \frac{\partial L(t,\tau)}{\partial[\hat{s}_i^{[l]}(t)]}h^{(1)}(\hat{u}_i^{[l]}(t)) \cdot \frac{\partial^\alpha[\hat{u}_i^{[l]}(t)]}{\partial[w_{ij}^{[l]}(t)]^\alpha} \tag{13}$$

with $i = 1, 2, \dots, n^{[l]}, j = 1, 2, \dots, n^{[l-1]}$. From (10), one has $\partial\hat{u}_i^{[l]}(t)/\partial w_{ij}^{[l]}(t) = \hat{s}_j^{[l-1]}(t)$. Then, refer to (8) and (9), the fractional-order gradients in (13) can be determined as

$$\frac{\partial^\alpha[\hat{u}_i^{[l]}(t)]}{\partial[w_{ij}^{[l]}(t)]^\alpha} = \frac{\hat{s}_j^{[l-1]}(t)}{\Gamma(2-\alpha)}\big|w_{ij,(k)}^{[l]}(t) - w_{ij,(k-1)}^{[l]}(t)$$
$$+ \varepsilon\big|^{1-\alpha} \tag{14}$$

where $w_{ij,(k)}^{[l]}(t)$ is the value of $w_{ij}^{[l]}(t)$ at the $k^{\text{th}}$ iteration, and $\hat{s}_{j,(k-1)}^{[l-1]}(t)$ is the value of $\hat{s}_j^{[l-1]}(t)$ at the $(k-1)^{\text{th}}$ iteration. Moreover, as we mentioned before, $\varepsilon$ is a small positive value to avoid potential singularities in the numerical computations. Then, combining Equations (13) and (14) yields

$$\frac{\partial^\alpha L(t,\tau)}{\partial[w_{ij}^{[l]}(t)]^\alpha} = \frac{\partial L(t,\tau)}{\partial[\hat{s}_{i,(k-1)}^{[l]}(t)]}h^{(1)}(\hat{u}_{i,(k-1)}^{[l]}(t))$$
$$\cdot \frac{\hat{s}_{j,(k-1)}^{[l-1]}(t)}{\Gamma(2-\alpha)}\big|w_{ij,(k)}^{[l]}(t) - w_{ij,(k-1)}^{[l]}(t)$$
$$+ \varepsilon\big|^{1-\alpha} \tag{15}$$

where $\frac{\partial L(t,\tau)}{\partial[\hat{s}_{i,(k-1)}^{[l]}(t)]}$ and $\hat{u}_{i,(k-1)}^{[l]}(t)$ are the values of $\frac{\partial L(t,\tau)}{\partial[\hat{s}_i^{[l]}(t)]}$ and $\hat{u}_i^{[l]}(t)$ at the $(k-1)^{\text{th}}$ iteration, respectively.

Particularly, in the context of image classification tasks, as depicted in Fig. 4, the neurons within the input layer produce Poisson spikes with an arrival rate following

Poisson's distribution, which is directly proportional to the pixel intensity. The input signal is then transformed into spatiotemporally defined spike events. Simultaneously, the loss function $L(t, \tau)$ for image classification tasks can be defined as

$$L(t, \tau) = \beta \sum_{i=1}^{n^{[o]}} (r_i^{[o]}(t) - \hat{s}_i^{[o]}(t))^2 \tag{16}$$

where $\beta$ is a weight factor associated with the loss term for the target neuron, $[o]$ signifies the output layer of the SNN, and $n^{[o]}$ represents the total number of output units. Additionally, $r_i^{[o]}(t)$ denotes the teaching signal in the form of spikes for the $i^{\text{th}}$ neuron within the output layer. It is important to note that we consider $r_i^{[o]}(t) = 1$ for the target neuron (class) and $r_i^{[o]}(t) = 0$ for non-target neurons. By further substituting Equation (16) into Equation (15), the loss gradients can be obtained as

$$\frac{\partial^\alpha L(t, \tau)}{\partial [w_{ij}^{[l]}(t)]^\alpha} = \xi_{i,(k-1)}^{[l]}(t) \frac{\hat{s}_{j,(k-1)}^{[l-1]}(t)}{\Gamma(2-\alpha)} \Big| w_{ij,(k)}^{[l]}(t) \\ - w_{ij,(k-1)}^{[l]}(t) + \varepsilon \Big|^{1-\alpha} \tag{17}$$

wherein the output layer ($l = o$), and hidden layer ($l = 1, 2, \ldots, o-1$) surrogate variables that reinforce the spike-timing dependency of synaptic weights are elucidated as

$$\xi_{i,(k-1)}^{[o]}(t) = -2\gamma\beta \cdot \left( r_i^{[o]}(t) - \hat{s}_{i,(k-1)}^{[o]}(t) \right) \\ \cdot \left[\!\left[ \left( \sum_{r=t-\tau+1}^{t} s_{i,(k-1)}^{[o]}(r) \geq 1 \right) ? 1 : 0 \right]\!\right] \\ \xi_{i,(k-1)}^{[l]}(t) = \gamma \cdot \sum_{h=1}^{n^{[l+1]}} \xi_{h,(k-1)}^{[l+1]}(t) \cdot w_{hi,(k-1)}^{[l+1]}(t) \\ \cdot \left[\!\left[ \left( \sum_{r=t-\tau+1}^{t} s_{i,(k-1)}^{[l]}(r) \geq 1 \right) ? 1 : 0 \right]\!\right] \tag{18}$$

In these equations, the $[\![ \text{condition} ? \text{rst1} : \text{rst2} ]\!]$ notation represents a conditional ternary operator, yielding "rst1" when the condition is true, and "rst2" otherwise. Additionally, $\gamma$ is defined as $\gamma = \theta^{-1}$, serving as the gain parameter introduced in Equation (5). Consequently, referring to (8), the synaptic weights are updated according to

$$w_{ij,(k+1)}^{[l]}(t) = w_{ij,(k)}^{[l]}(t) - \mu \frac{\partial^\alpha L(t, \tau)}{\partial [w_{ij}^{[l]}(t)]^\alpha} \tag{19}$$

where $\mu$ represents the learning rate, and the loss gradient is determined by Equation (15), or, in the context of image-related tasks, Equation (17).

**Theorem 2**. The synaptic weights of a fully-connected feedforward multi-layer SNN, considering the updating rule for the fractional loss gradient as presented in Equation (15),

---

**Algorithm 1** (FO-STDGD) Supervised learning of a fully-connected feedforward multi-layer SNN through a fractional-order spike-timing-dependent gradient descent method.

1: **for** $s = 1, 2, \ldots, N$ **do** ($N$ is the number of training samples)
2:   **for** $t = 1, 2, \ldots, T$ **do**
3:     **for** $l = 1, 2, \ldots, o$ **do**
4:       $u^{[l]}(t) = u^{[l]}(t-1) + w^{[l]}(t)s^{[l-1]}(t)$
5:       **for** $i = 1, 2, \ldots n^{[l]}$ **do**
6:         **if** $u_i^{[l]}(t) > \theta$ **then**
7:           $s_i^{[l]} = 1$ and $u_i^{[l]}(t) = u_{reset}$
8:     **if** $t\%\tau == 0$ **then**
9:       **for** $l = o, o-1, \ldots, 2$ **do**
10:         **for** $i = 1, 2, \ldots, n^{[l]}$ **do**
11:           $\hat{s}_i^{[l]} = \frac{1}{\tau} \sum_{s=1}^{\tau} s_i^{[l]}(t-s+1)$
12:           **for** $j = 1, 2, \ldots, n^{[l-1]}$ **do**
13:             Compute $\frac{\partial^\alpha L(t, \tau)}{\partial [w_{ij}^{[l]}(t)]^\alpha}$ from Equation (15) or (17)
14:             Update $\frac{\partial L(t, \tau)}{\partial [s_j^{[l-1]}(t)]}$ from Equation (12) or (18)
15:             $w_{ij,(k-1)}^{[l]}(t) \leftarrow w_{ij,(k)}^{[l]}(t)$
16:             $w_{ij,(k+1)}^{[l]}(t) = w_{ij,(k)}^{[l]}(t) - \mu \frac{\partial^\alpha L(t, \tau)}{\partial [w_{ij}^{[l]}(t)]^\alpha}$
17:             $w_{ij,(k)}^{[l]}(t) \leftarrow w_{ij,(k+1)}^{[l]}(t)$
18: **return** $w_{ij}^{[l]}(t)$

---

if updated by Equation (19), converge to the real extreme point $w_{ij}^{[l]*}(t)$ that minimizes the loss function.

**Proof**. The following proof is given by contradiction. Assume that $w_{ij}^{[l]}(t)$ converges to a point $w_{ij}^{[l]'}(t)$ that is different from $w_{ij}^{[l]*}(t)$, i.e., $\lim_{k\to\infty} w_{ij,(k)}^{[l]}(t) = w_{ij}^{[l]'}(t) \neq w_{ij}^{[l]*}(t)$ pointwisely for fixed $t$, we know that for any $\varepsilon > 0$, there exists $N > 0$ such that $\left| w_{ij,(k-1)}^{[l]}(t) - w_{ij}^{[l]'}(t) \right| < \varepsilon < \left| w_{ij}^{[l]*}(t) - w_{ij}^{[l]'}(t) \right|$ if $k - 1 > N$. The first order necessary condition for optimality implies that $\frac{\partial L(t, \tau)}{\partial w_{ij}^{[l]*}(t)} = 0$, thus we can choose a small positive $\delta$ that satisfies $\delta = \inf_{k-1>N} \left| \frac{\partial L(t, \tau)}{\partial w_{ij,(k-1)}^{[l]}(t)} \right| > 0$. Subsequently, it follows from Equation (9) or Equation (19) that (we omit the fixed $t$ in the notations of subsequent derivations)

$$\left| w_{ij,(k+1)}^{[l]} - w_{ij,(k)}^{[l]} \right| \\ = \left| \frac{\mu}{\Gamma(2-\alpha)} \cdot \frac{\partial L}{\partial w_{ij,(k-1)}^{[l]}} \cdot \left| w_{ij,(k)}^{[l]} - w_{ij,(k-1)}^{[l]} \right|^{1-\alpha} \right| \\ = \frac{\mu}{\Gamma(2-\alpha)} \cdot \left| \frac{\partial L}{\partial w_{ij,(k-1)}^{[l]}} \right| \cdot \left| w_{ij,(k)}^{[l]} - w_{ij,(k-1)}^{[l]} \right|^{1-\alpha} \\ \geq \lambda \left| w_{ij,(k)}^{[l]} - w_{ij,(k-1)}^{[l]} \right|^{1-\alpha} \tag{20}$$

with $\lambda = \frac{\mu\delta}{\Gamma(2-\alpha)} > 0$.

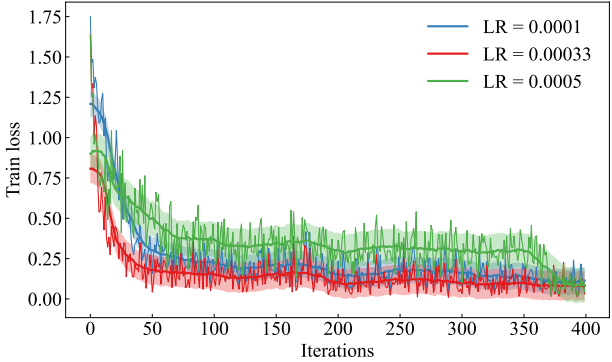

Fig. 5. Batch training loss of the three-layer SNN over 400 iterations with three different learning rate.

Meanwhile, for a sufficient small $\varepsilon$ that satisfies $2\varepsilon < \lambda^{1/\alpha}$ , the convergence of $w_{ij,(k)}^{[l]}$ and the triangular inequality gives following relation.

$$
\begin{aligned}
\left| w_{ij,(k)}^{[l]} - w_{ij,(k-1)}^{[l]} \right| & \\
\leq \left| w_{ij,(k)}^{[l]} - w_{ij}^{[l]'} \right| & + \left| w_{ij,(k-1)}^{[l]} - w_{ij}^{[l]'} \right| \\
< 2\varepsilon & < \lambda^{1/\alpha}
\end{aligned}
\tag{21}
$$

Combining inequalities in Equations (20) and (21) yields

$$
\left| w_{ij,(k+1)}^{[l]} - w_{ij,(k)}^{[l]} \right| > \left| w_{ij,(k)}^{[l]} - w_{ij,(k-1)}^{[l]} \right|
\tag{22}
$$

The inequality expressed in Equation (22) signifies that the sequence $\left\{ w_{ij,(k)}^{[l]} \right\}_{k=1}^{\infty}$ does not adhere to Cauchy criterion. In a complete vector space, such as $\mathbb{R}$, a non-Cauchy sequence is inherently non-convergent. This incongruity contradicts our initial assumption that $\lim_{k \to \infty} w_{ij,(k)}^{[l]}(t) = w_{ij}^{[l]'}(t) \neq w_{ij}^{[l]*}(t)$. □

**Remark 2.** Theorem 2 substantiates the convergence of synaptic weights towards the real minimum of the loss function when the update rule (19) is applied. This theoretical assurance promises the establishment of a novel supervised learning model, termed the Fractional-Order Spike Timing Dependent Gradient Descent (FO-STDGD), which is comprehensively delineated as Algorithm 1. The incorporation of a fractional-order loss gradient in the backpropagation process of multilayer SNN, along with the anticipated enhanced convergence efficacy as indicated by the fractional gradient descent method (8) and illustrated in Fig. 2, endows our FO-STDGD with the qualities of potentially expedited convergence rate and higher learning precision in comparison to traditional integer-order learning models.

## 3. Experiments and discussions

To assess the FO-STDGD learning model proposed in this work, we initially apply it to the well-established benchmark task of recognizing handwritten digits from the MNIST dataset, as outlined in Section 3.1. The MNIST dataset is a time-honored benchmark within the machine learning community, offering a proven ground for validating our theoretical results. In section 3.2. and 3.3., we delve deeper into comparing the classification accuracy and computational cost of several typical learning algorithms in the identification of the MNIST dataset and a more SNN-aligned DVS128 Gesture dataset [52]. This extension is critical for gauging the flexibility and resilience of the FO-STDGD learning model when applied to a spectrum of datasets that vary in complexity and structure.

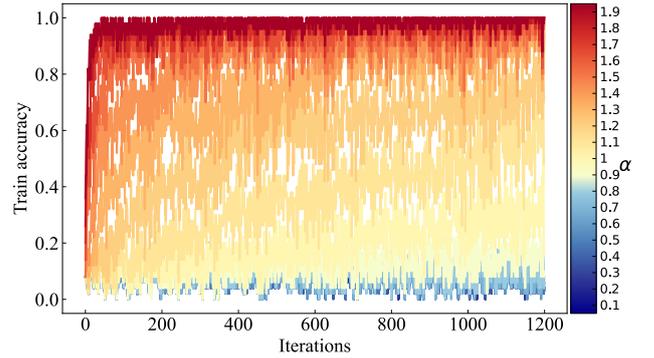

Fig. 6. Batch training accuracy of the three-layer SNN over 1200 iterations (one epoch) with different fractional gradient orders $\alpha$.

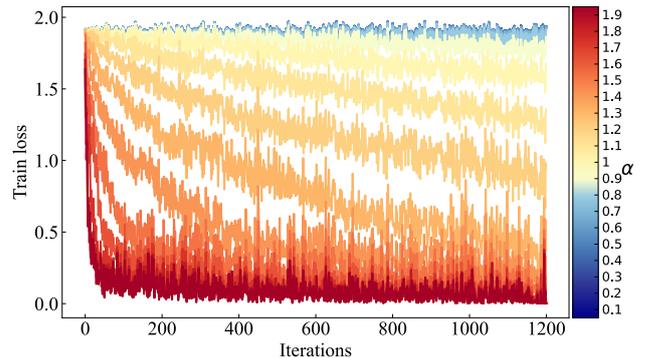

Fig. 7. Batch training loss of the three-layer SNN over 1200 iterations (one epoch) with different fractional gradient orders $\alpha$.

**Table 1**
Parameters used in FO-STDGD training.

| Parameter | Value | Parameter | Value |
|---|---|---|---|
| Training window $T$ | 100 ms | Learning rate $\mu$ | 0.00033 |
| Average time $\tau$ | 50 ms | Loss weight $\beta$ | 2.0 |
| Threshold voltage $\theta$ | 15 V | Batch size $m$ | 50 |
| Error parameter $\varepsilon$ | $1e-5$ | Num. of Iteration | 1200 |

### 3.1. Benchmark problem simulation and analysis

The MNIST set consists of 60,000 grayscale handwritten digits for training and another 10,000 images for testing. As displayed in Fig. 4, we first convert the $28 \times 28$ pixel intensity matrix per image into spike trains released from 784 input neurons, and the input spike trains are generated by emulating a Poisson process with its spike arrival (firing) rate proportional to the normalized pixel intensity. The full structure of the deep SNN consists of an input layer with 784 neurons, the first hidden layers with 500 neurons, the second hidden layer with 150 neurons and an output layer with 10 neurons. Each neuron in the output layer corresponds to a class label ( $0 \sim 9$ ) in MNIST set. The corresponding hyperparameters in the training procedures are set by grid search. For instance, Fig. 5 compares the learning process of the three-layer SNN over 400 iterations under three close but different learning rates. For each iteration, a batch of sample digits is scanned sequentially, with its batch size $m = 50$.

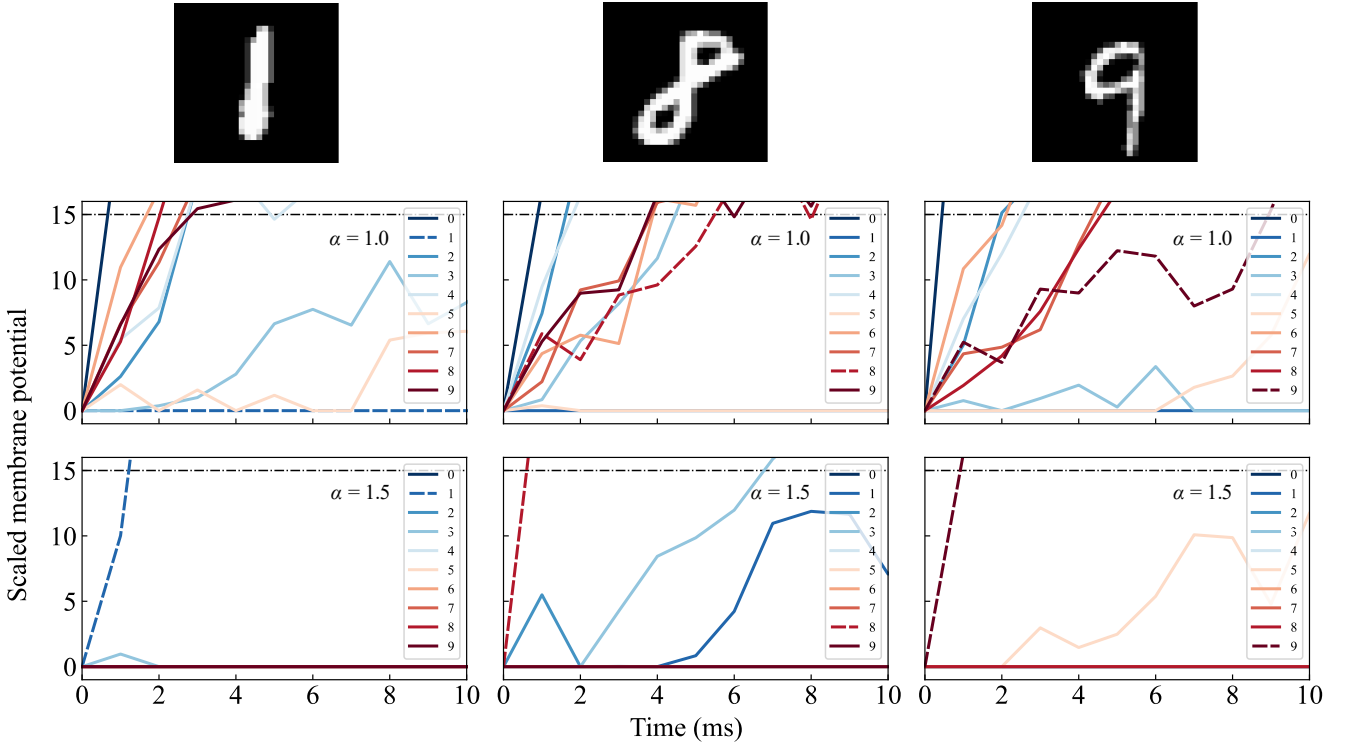

Fig. 8. Two sets of temporally evolved membrane potentials of output neurons in response to three randomly selected handwritten digits ('1', '8' and '9'), with the first and second sets of potentials generated by FO-STDGD with $\alpha = 1.0$ and $\alpha = 1.5$, respectively.

**Table 2**
Average training, testing accuracy and training loss

| Order ($\alpha$) | Training Accuracy | Testing Accuracy | Training Loss |
|---|---|---|---|
| 0.1 | 0.0498 | 0.0466 | 2.8299 |
| 0.2 | 0.0497 | 0.0466 | 2.8298 |
| 0.3 | 0.0507 | 0.0462 | 2.8297 |
| 0.4 | 0.0501 | 0.0467 | 2.8294 |
| 0.5 | 0.0507 | 0.0474 | 2.8283 |
| 0.6 | 0.0520 | 0.0491 | 2.8247 |
| 0.7 | 0.0564 | 0.0534 | 2.8133 |
| 0.8 | 0.0734 | 0.0697 | 2.7775 |
| 0.9 | 0.1331 | 0.1318 | 2.6640 |
| 1.0 | 0.3770 | 0.3837 | 2.3750 |
| 1.1 | 0.6758 | 0.6679 | 2.0155 |
| 1.2 | 0.7990 | 0.7947 | 1.5729 |
| 1.3 | 0.8802 | 0.8704 | 1.1116 |
| 1.4 | 0.9121 | 0.9060 | 0.7581 |
| 1.5 | 0.9419 | 0.9338 | 0.5008 |
| 1.6 | 0.9577 | 0.9576 | 0.3310 |
| 1.7 | 0.9732 | 0.9713 | 0.2208 |
| 1.8 | 0.9796 | 0.9760 | 0.1640 |
| 1.9 | 0.9795 | 0.9764 | 0.1454 |

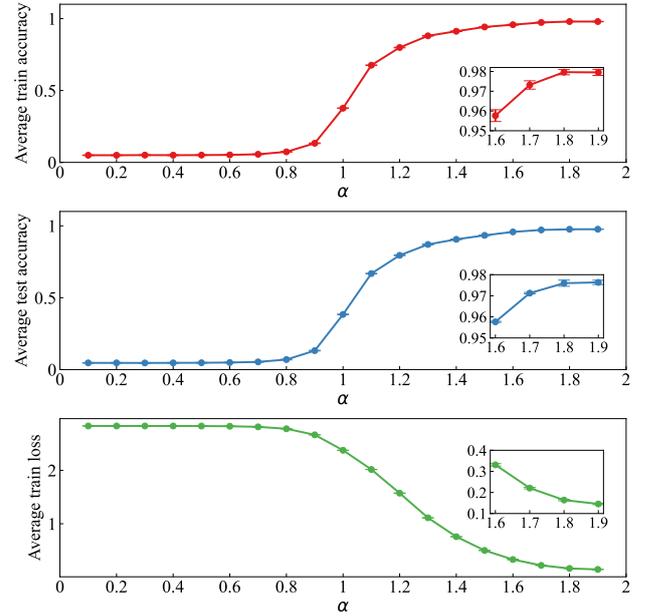

Fig. 9. Average training, testing accuracy and training loss for different fractional gradient orders $\alpha$ in two epochs of training, with error bar denoting the standard deviation after ten trials of experiments.

The training process for 20,000 samples in Fig. 5 suggests that a learning rate $\mu$ of 0.00033 is superior to its nearby values. All the parameters used in FO-STDGD training are listed in Table 1.

The experiments are carried out for 19 different $\alpha$ values, respectively. In the simulations, all initial values for the

**Table 3**
Performance comparison among different learning algorithms for SNNs on the entire MNIST dataset.

| Model | Network Structure | Learning Paradigm | Coding Scheme | Epochs | Classification Accuracy |
|---|---|---|---|---|---|
| Diehl et al. [55] | 784-6400 | Unsupervised | Temporal-based | HI | 0.95 |
| Tavanaei et al. [59] | Spiking CNN | Unsupervised | Temporal-based | HI | 0.9836 |
| Lee et al. [35] | 784-500-500-10 | Supervised | Rate-based | 200 | 0.987 |
| Rueckauer et al. [57] | 7-layer SNN | Supervised | Temporal-based | HI | 0.9944 |
| Tavanaei et al. [46] | 784-500-150-10 | Supervised | Rate-based | HI | 0.972 |
| Kheradpisheh et al. [60] | Spiking CNN | Unsupervised | Temporal-based | HI | 0.984 |
| Mostafa [41] | 784-400-400-10 | Supervised | Temporal-based | 100 | 0.9692 |
| Mostafa [41] | 784-800-10 | Supervised | Temporal-based | 100 | 0.972 |
| O'Connor et al. [56] | 784-500-10 | Supervised | Temporal-based | 2 (25) | 0.965 (0.9763) |
| Shrestha et al. [58] | Spiking CNN | Supervised | Temporal-based | 1000 | 0.9936 |
| Wu et al. [36] | 784-400-10 | Supervised | Temporal-based | 2 (200) | 0.94 (0.9848) |
| Shen et al. [54] | 784-800-800-10 | Supervised | Temporal-based | 2 (10) | 0.9685 (0.9784) |
| Comsa et al. [34] | 784-340-10 | Supervised | Temporal-based | 1000 | 0.9796 |
| Zhang et al. [44] | 784-800-10 | Supervised | Temporal-based | 150 | 0.985 |
| **Our FO-STDGD** | 784-500-150-10 | Supervised | Temporal-based | 2 (10) | 0.9764 (0.9843) |
| **Our FO-STDGD** | 784-400-10 | Supervised | Temporal-based | 2 (10) | 0.9746 (0.9831) |
| **Our FO-STDGD** | 784-700-10 | Supervised | Temporal-based | 2 (10) | 0.9761 (0.9856) |
| **Our FO-STDGD** | 784-1000-10 | Supervised | Temporal-based | 2 (10) | 0.9775 (0.987) |

Note: HI in column *Epochs* stands for "high", the number in the parenthesis indicates the boosted number of epochs, e.g., in O'Connor et al. [56], the best classification accuracy is 0.965 for only two running epochs, and the best accuracy is boosted to 0.9763 with 25 running epochs.

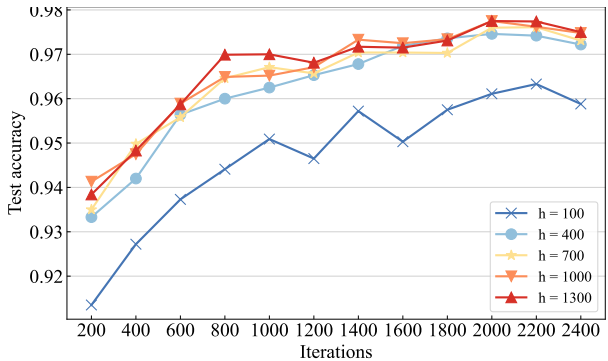

Fig. 10. Classification accuracy exhibited by a two-layer SNN with different number of hidden layer IF neurons along the training process.

weight matrix $w_{ij}^{[l]}$ are required to be randomly selected from the standard normal distribution. Since the values of hyperparameters are invariant to $\alpha$, we can evaluate the performance of FO-STDGD with different $\alpha$ values. Fig. 6 and Fig. 7 show the batch training accuracy and batch training loss during the training process of the SNN in one epoch with $\alpha$ varying from 0.1 to 1.9. It can be observed that the convergence speed of the FO-STDGD algorithm increases as the value of $\alpha$ increases, and the average training accuracy of FO-STDGD with a larger $\alpha$ value is also higher for the same number of iterations. This observation is consistent with the evolving trend of the convergence curves observed in Fig. 2, validating our proposal that the learning performance can be improved by manipulating the fractional order of FO-STDGD.

Faster convergence rate is associated with larger $\alpha$ because of the augmented long-term potentiation (LTP) for the synapses between target and hidden neurons. Fig. 8 displays the first-tenth millisecond evolution of the membrane potential for ten output neurons classifying three randomly selected handwritten digits (1,8 and 9) with two different $\alpha$ values (upper row $\alpha = 1.0$, lower row $\alpha = 1.5$). The study reveals that elevated $\alpha$ values correlate with a quicker rise in the membrane potential of target neurons to threshold levels, thereby accelerating spike initiation relative to non-target neurons. This acceleration yields a decrease in the neuronal system's response latency by over 5 milliseconds.

To evaluate the generalizability of the training model and demonstrate a performance comparison with other training approaches, we increased the training session to two epochs ($2 \times 1200$ iterations) and repeated the experiment ten times, each time randomly reselecting the initial values of the synaptic weights and randomly reshuffling the order of training samples. As a result, the average training accuracy, testing accuracy and average training loss at different $\alpha$ values are presented in Table 2, with the corresponding standard deviation denoted by error bars in Fig. 9. It is shown that the distribution of the testing accuracy from repeated experiments has a very small standard deviation ($\sigma(\alpha = 1.9) \approx 0.00102$), which suggests an excellent generalizability of our training model.

In addition to the three-layer SNN structure (784-500-150-10), the training experiments are also carried out on a two-layer SNN with its number of hidden neurons varying from 100 to 1300 and the fractional order $\alpha$ fixed to 1.9. Fig. 10 displays the classification accuracy of those distinct two-layer SNNs along the training process. In addition to the general pattern of elevated classification accuracy as the number of hidden neurons and training iterations increases, we also find that for SNNs with 1000 and 1300 hidden neurons, the best accuracy of 0.9775 is achieved with 2000 iterations if training is run for only two epochs. It suggests that single-hidden layer SNNs can already show competitive classification accuracy with sufficiently large number of hidden neurons (e.g., 1000), and there is no need to pursue a larger size of network structure.

### 3.2. Comparison to other models and performance analysis

Performance evaluation and comparison of classification models requires consideration of the training strategies employed, such as the network structure, the number of training epochs, and the spike coding schemes (rate-based or temporal-based). The impact of spike coding schemes on classification performance is omitted in this study because training models with training latency (i.e., the time duration to train SNN for each input sample) greater than 40 milliseconds (~70 ms for our simulated SNNs) have been reported to have similar classification accuracies with different spike coding schemes [53]. We thus emphasize two other features of the training models (i.e., network structure and number of epochs) and report our classification accuracy compared to other typical training approaches in Table 3. Specifically, our three-layer SNN network structure consists of two hidden layers each containing 500 and 150 neurons, which is the same network structure used in Tavanaei's BP-STDP [46], and is simpler than the network structures in Lee's backpropagated SNN [35], Mostafa's temporally coded feedforward SNN [41] and Shen's HybridSNN [54]. We find that our FO-STDGD can achieve an accuracy of 0.9764 with two training epochs and boost the accuracy to 0.9843 with ten training epochs. It is reasonable to observe such an improvement in our classification accuracy with the increase of training epochs because most of the wrongly classified samples in the early stages of training can adaptively readjust the synaptic weights of some over-activated excitatory hidden neurons in their later training epochs. Our spike-timing dependent self-tunned synapses can then achieve a similar effect as the dropout regularization technique in non-spiking neural networks and avoid the overfitting appeared in the early training stages. The self-regularization of our trained SNNs can also be observed in Fig. 10, where the rebound of test accuracy during training is gradually weakened as iteration number increases. And favorably, it is found that our three-layer SNN achieves the best classification accuracy among all the three-layer SNNs in Table 3.

Furthermore, we demonstrate a comparison of our FO-STDGD with other learning approaches in terms of two-layer SNN architectures. It is observed that the performance of our FO-STDGD is much improved compared to the traditional unsupervised learning model (i.e., STDP), as we can identify up to 370 more images (out of 10,000 images in test set) than the STDP scheme [55]. In addition, we find that our model with 400 hidden neurons can achieve an accuracy of 0.9746 (0.9831) within two (ten) training epochs. This result indicates the superiority of our learning model over Comsa's alpha synaptic function approach [34], O'Connor's equilibrium propagation (Eq-Prop) method [56], and Wu's spatio-temporal backpropagation (STBP) approach [36]. It should be noted that Wu's STBP achieved their best accuracy of 0.9848 by introducing a convolutional layer and trained their model for 200 epochs. Meanwhile, we discover that the two-layer SNNs with 700 and 1000 hidden neurons trained by our FO-STDGD also outperform the SNNs with similar network structures presented in other's research, such as Mostafa's SNN with 800 hidden neurons and Zhang's SNN with 800 hidden neurons [41,44]. However, we should mention that we did not evaluate the performance of FO-STDGD under much deeper SNNs or spiking convolutional neural networks (SCNNs), because in such cases there is a higher requirement for hardware computing power and hyperparameter conditioning.

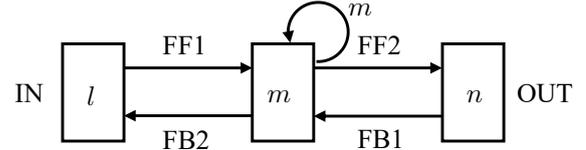

| Algorithm | FF1 | FF2 | FB1 | FB2 |
|-----------|-----|-----|-----|-----|
| STDP | $\mathcal{O}(lm)$ | $\mathcal{O}(mn)$ | - - - - | - - - - |
| STBP | $\mathcal{O}(lm)$ | $\mathcal{O}(mn)$ | $\mathcal{O}(nm)$ | $\mathcal{O}(ml)$ |
| Eq-Prop | $\mathcal{O}(lm)$ | $\mathcal{O}(mn)$ | $\mathcal{O}(nm)$ | $\mathcal{O}(ml)$ |
| BP-STDP | $\mathcal{O}(lm)$ | $\mathcal{O}(mn)$ | $\mathcal{O}(nm)$ | $\mathcal{O}(ml)$ |
| FO-STDGD | $\mathcal{O}(lm)$ | $\mathcal{O}(mn)$ | $\mathcal{O}(nm)$ | $\mathcal{O}(ml)$ |

Fig. 11. Overview of algorithmic complexity, $\mathcal{O}(\cdot)$, per epoch within the learning phase, encompassing both feedforward (FF) and feedback (FB) propagation processes. The variables $l$, $m$, and $n$ represent the neuron count in the input, hidden, and output layers of the network, respectively. Algorithms evaluated include STDP [55], STBP [36], Eq-Prop [56], BP-STDP [46], and our FO-STDGD.

In Table 3, we list several training models under the use of more complex network structures. For example, Rueckauer proposed a seven-layer spiking SNN to achieve a spectacularly high accuracy of 0.9944 and Shrestha achieved an accuracy of 0.9936 with a SCNN (28x28-12c5-2a-64c5-2a-10o, represents a 6-layer SNN with 28×28 input, followed by 12 convolution filters (5×5), followed by 2×2 aggregation layer, followed by 64 convolution filters (5×5), followed by 2×2 aggregation layer, and finally a dense layer connected to 10 output neurons) [57,58]. Two other impressing SCNNs were proposed by Tavanaei [59] and Kheradpisheh [60], where they achieved high accuracy of 0.9836 and 0.984, respectively. While the performance of our FO-STDGD has not been evaluated on deeper SNN architectures or SCNNs, it is essential to acknowledge the algorithm's robust adaptability. The architectural versatility of FO-STDGD is one of its standout features; the algorithm's design is fundamentally agnostic to network configurations, which enables it to adjust to the diverse and complex architectures inherent to both SCNNs and spiking recurrent neural networks (SRNNs) with efficiency.

### 3.3. Computational cost analysis

In evaluating the performance of learning algorithms, computational cost emerges as a crucial metric. This metric, denoted as $C_i$ for the $i$-th algorithm, is quantified by

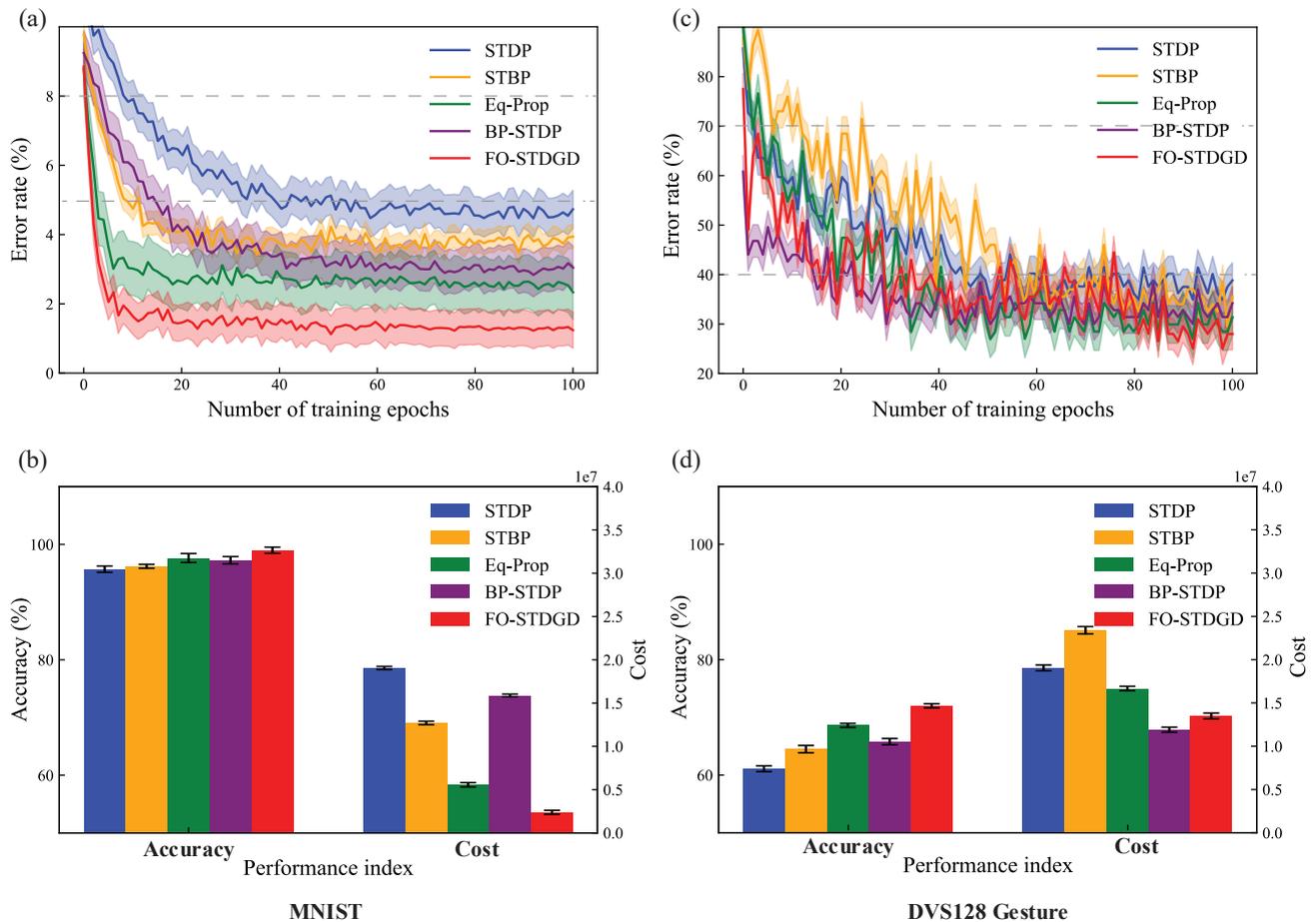

**MNIST**  **DVS128 Gesture**

Fig. 12. Evaluation of five algorithms for digit recognition in the MNIST dataset and the DVS128 Gesture dataset using an SNN architecture with a 784-1000-10 configuration. (a) and (c) illustrate the progression of test set accuracy, demarcated by upper (8%, 70%) and lower (5%, 40%) bounds that govern the associated computational costs. (b) and (d) present aggregated histograms detailing the peak average accuracy and the computational costs, along with their standard deviation values calculated from five repetitive runs of simulations.

multiplying the mean epochs required to reach a specific error threshold with the algorithm's epoch-wise complexity, $\mathcal{O}(n)_i$. The per-epoch algorithmic complexity for our five evaluated algorithms are summarized in Fig. 11. Formally, $C_i$ is computed as [61]

$$C_i = \frac{1}{N} \sum_{j=1}^{N} \underset{x}{\mathrm{argmin}}(f_i(x) = err_j) \cdot \mathcal{O}(n)_i \tag{23}$$

where $\mathrm{argmin}(\cdot)$ identifies the epoch at which the minimum error level is first achieved, $f_i(x)$ delineates the error rate trajectory over epochs, $\mathcal{O}(n)_i$ represents the algorithmic complexity in relation to the number of parameters $n$, and $N$ signifies the total number of discrete error levels under consideration. By incorporating this metric, we can assess algorithms not just on their accuracy or convergence rate but also on their computational demands, thereby providing a comprehensive analysis of their practical applicability in various settings.

The comparative analysis delineated in Fig. 12(a) and (b) reveals that the proposed FO-STDGD outstrips competing schemes in terms of both accuracy and computational efficiency for identification of MNIST dataset. Compared to the other four algorithms, FO-STDGD not only attains the zenith of accuracy, surpassing 98.7%, but also showcases a significant decrease in computational costs by a minimum of 60%, a testament to its optimized design. The juxtaposition of FO-STDGD with other contenders such as BP-STDP and Eq-Prop shows that while these algorithms maintain competitive accuracy, they fall behind in computational economy. The STDP and STBP algorithms, while offering moderate accuracy, incurs a disproportionate computational cost, which could be prohibitive in scenarios with limited resources.

Furthermore, we extend our analysis to the DVS128 Gesture dataset. As elucidated in Figures 12(c) and (d), the FO-STDGD model secures a commendable accuracy of 72%, which notably surpasses the 68.5% achieved by the subsequent Eq-Prop model. Additionally, the FO-STDGD maintains competitive computational efficiency, exhibiting only a marginal increase in cost relative to the BP-STDP algorithm. In essence, the FO-STDGD algorithm's superior performance metrics underscore its potential as the preferred choice for tasks where both precision and cost-effectiveness are paramount. This superior amalgamation of high accuracy with minimized computational cost underscores the algorithm's innovative edge and reinforces its suitability for extensive application in real-world computational environments where efficiency is as critical as performance.

## 4. Conclusion

In this paper, a spatiotemporally encoded supervised learning model, namely the fractional-order spike-timing-dependent gradient descent (FO-STDGD) algorithm, was developed for deep SNNs. We first derived and verified a nonlinear activation function that correlates the quasi-instantaneous firing rate and the temporal membrane potential of nonleaky IF neurons. The FO-STDGD scheme can adjust its training convergence rate by freely selecting the order of the fractional gradient between 0 and 2. We tested our learning scheme on the MNIST dataset and found that as the order of FO-STDGD increased from 0.1 to 1.9, the classification accuracy and convergence rate improved significantly. In addition, it was demonstrated that our scheme compares favorably with other training approaches as we can achieve higher accuracy (0.9843 for the three-layer SNN and 0.987 for the two-layer SNN) with less training epochs and simpler network structure. Upon evaluation using the DVS128 Gesture dataset, our FO-STDGD model attained an accuracy of 72%, which notably surpasses the performance of four other algorithms.

The computational cost analysis further solidified FO-STDGD's standing as an algorithm of high efficacy. It demonstrated an admirable balance of computational thrift and training performance, wherein the algorithm not only improved in accuracy with an increasing order but also did so with a noticeable reduction in the number of training epochs required. Such an outcome highlights the algorithm's proficiency in utilizing computational resources prudently while enhancing learning efficacy. This duality of benefits, when considered alongside the spike-timing dependence of the update rules and the fractional-order gradient's nonlocal influence, positions FO-STDGD as an optimal candidate for future neuromorphic applications. And these promising findings suggest a potential high-efficiency and low-power neuromorphic (hardware) implementation of deep SNNs in future work, paving the way for advancements in both algorithmic development and neuromorphic engineering.

## Acknowledgments

The work was partially sponsored by Office of Naval Research Young Investigator Program, Award No.: N00014-21-1-2585, and National Science Foundation, Grant CNS-1726865.

## Author biography

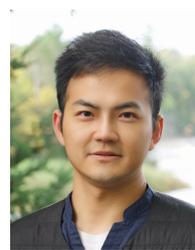

**Yi Yang** received the B.Eng. degree in Thermal Energy and Power Engineering from Huazhong University of Science and Technology, Wuhan, China, in 2015, and the M.Sc. in Mechanical Engineering from the University of Michigan, Ann Arbor, U.S.A., in 2017, and the Ph.D. degree in Engineering Technology from Purdue University, West Lafayette, U.S.A., in 2021. From 2021 to 2023, he was associated with the School of Engineering Technology, Purdue University, as a post-doctoral researcher. He is currently a post-doctoral research fellow with the Multi-Scale Medical Robotics Center, and the Department of Mechanical and Automation Engineering, The Chinese University of Hong Kong, Hong Kong, China.

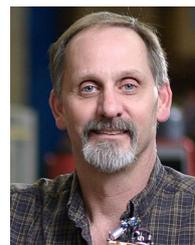

**Richard M. Voyles** received the B.S. degree in electrical engineering from Purdue University, West Lafayette, IN, USA, in 1983, the M.S. degree in mechanical engineering from Stanford University, Stanford, CA, USA, in 1989, and the Ph.D. degree in robotics from Carnegie Mellon University, Pittsburgh, PA, USA, in 1997. He is the Daniel C. Lewis Professor of the Polytechnic and a University Faculty Scholar with Purdue University. He is the Founder of the Collaborative Robotics Lab, the Director of the Purdue Robotics Accelerator, and the Site Director of the NSF Center for Robotics and Sensors for Human Well-Being (RoSe-HUB). His research interests include miniature, constrained robots, mobile manipulation, multirobot coordination, programming by human demonstration, and haptic sensors and actuators.

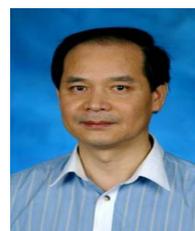

**Haiyan Henry Zhang** is a Full Professor of Engineering Technology and the founding director of Center for Technology Development at Purdue University. He received his Ph.D. degree from University of Michigan-Ann Arbor in 1996. With the multidisciplinary engineering background and industrial experience, his research foci are analytical mechatronic design, advanced manufacturing control, and design of traditional or hybrid electric vehicle powertrains.

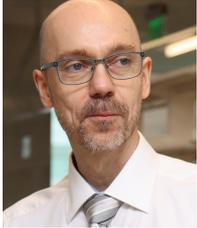 **Robert A Nawrocki** is an Assistant Professor in the School of Engineering Technology (SOET) at Purdue University. His current research interests include physically flexible organic electronics with the application in biopotential monitoring and soft robotics, as well as neuromorphic systems, chemical sensors, smart (meta) materials and neuroscience. He completed his BS at New Jersey Institute of Technology, PhD at University of Denver, his internship at Eidgenössische Technische Hochschule Zürich, Switzerland (ETH), and his postdoctoral research at University of Colorado Boulder, University of Nova Gorica, Slovenia, and The University of Tokyo, Japan.